\documentclass{article}

\twocolumn

\RequirePackage{times}

\RequirePackage{fancyhdr}
\RequirePackage{xcolor} % changed from color to xcolor (2021/11/24)
\RequirePackage{algorithm}
\RequirePackage{algorithmic}
\RequirePackage{natbib}
\RequirePackage{eso-pic} % used by \AddToShipoutPicture
\RequirePackage{forloop}
\RequirePackage{url}

\date{}

\usepackage{microtype}
\usepackage{graphicx}
\usepackage{subfigure}
\usepackage{booktabs} % for professional tables

\usepackage{hyperref}

\usepackage{amsmath}
\usepackage{amssymb}
\usepackage{mathtools}
\usepackage{amsthm}

\usepackage[capitalize,noabbrev]{cleveref}

\theoremstyle{plain}

\theoremstyle{definition}

\theoremstyle{remark}

\usepackage[textsize=tiny]{todonotes}

\newcommand{\popd}{\mathrm{pop}}

\newcommand{\gen}{\mathrm{gen}}

\DeclareMathOperator*{\argmin}{argmin}

\author{David McAllester \\ Toyota Technologicical Institute at Chicago (TTIC)}

\title{On the Mathematics of Diffusion Models}

\begin{document}

\maketitle

\begin{abstract}
  This paper gives direct derivations of the differential equations and likelihood formulas of diffusion models assuming only knowledge of Gaussian distributions.
  A VAE analysis derives both forward and backward stochastic differential equations (SDEs) as well as non-variational integral expressions
  for likelihood formulas.
  A score-matching analysis derives the reverse diffusion ordinary differential equation (ODE) and a family of reverse-diffusion SDEs parameterized by noise level.
  The paper presents the mathematics directly with attributions saved for a final section.
\end{abstract}

\section{The Diffusion Stochastic Differential Equation (SDE)}
\label{diffusion}

We assume a population density $\popd(y)$ for $y \in R^d$.  Of particular interest is a population of images.
For a given time resolution $\Delta t$ we consider the discrete-time process $z(0),z(\Delta t),z(2\Delta t),z(3\Delta t),\ldots$ defined by
\begin{eqnarray}
  z(0) & = & y,\;\;\;y \sim \popd(y) \nonumber \\
  \nonumber \\
  z(t+\Delta t) & = & z(t) + \sqrt{\Delta t}\; \epsilon,\;\;\;\epsilon \sim {\cal N}(0,I) \label{forward}
\end{eqnarray}
For $\epsilon \sim {\cal N}(0,I)$ each coordinate of the random variable $\sqrt{\Delta t} \;\epsilon$ has standard deviation $\sqrt{\Delta t}$ and variance  $\Delta t$.
A sum of Gaussians is a Gaussian with variance equal to the sum of the variances.  This implies that the variance of the coordinates of $z(n\Delta t)$ conditioned on $z(0)$
is $n\Delta t$.
Taking this to limit as $\Delta t \rightarrow 0$ we get that the variance of the coordinates of $z(t)$ given $z(0)$ is $t$.
More generally in the limit of $\Delta t \rightarrow 0$ equation (\ref{forward}) holds for sampling $z(t+ \Delta t)$ given $z(t)$ for any (continuous) non-negative $t$ and $\Delta t$.
This is the density (measure) on continuous functions $z(t)$ defined by the diffusion SDE (\ref{forward}).

\section{The VAE Analysis}

We first review general variational auto-encoders (VAEs).  Let $\popd(y)$ be any population distribution or density on any set ${\cal Y}$.
Let $p(z|y)$ be any distribution or density on any set ${\cal Z}$ conditioned on $y \in {\cal Y}$.  For the joint distribution defined by $\popd(y)$
and $p(z|y)$ we have the following for $p(y,z) > 0$ and where $p_\gen(z)$ and $p_\gen(y|z)$ are abitrary ``generator'' distributions or densities.
{\small
  \begin{eqnarray}
  \ln \popd(y) & = & \ln \frac{p(y)p(z|y)}{p(z|y)} \nonumber \\
  \nonumber \\
  & = & \ln \frac{p(z)p(y|z)}{p(z|y)} \nonumber \\
  \nonumber \\
  -\ln \popd(y) & = & E_{z|y}\left[\ln \frac{p(z|y)}{p(z)}\right] - E_{z|y}[\ln p(y|z)] \nonumber \\
  \nonumber \\
  & = & \left\{\begin{array}{l} KL(p(z|y),\;p(z)) \\ \\ - E_{z|y}\;[\ln p(y|z)]\end{array}\right. \label{VAE-kl}\\
  \nonumber \\
  H(\popd(y)) & \leq & \left\{\begin{array}{l}E_y\;\;KL(p(z|y),\;p_\gen(z)) \\ \\- E_{y,z}\;\ln p_\gen(y|z)\end{array}\right.\label{ELBO}
  \end{eqnarray}
  }Here (\ref{ELBO}) follows from (\ref{VAE-kl}) and the fact that cross-entropies upper bound entropies.  Typically the generator distributions are trained
by minimizing the ``variational bound'' in (\ref{ELBO}).  One also typically optimizes the encoder distribution or density $p(z|y)$.
However, for diffusion models the encoder is defined by the diffusion process.

The diffusion process can be analyzed as a Markovian VAE.  In a Markovian VAE we have
$z = z_1,\ldots z_N$ such that the distribution on $y,z_1,\ldots z_N$ is defined by $\popd(y)$, $p(z_1|y)$, and $p(z_{i+1}|z_i)$.
For a Markovian VAE we have the following for $p(y,z_1,\ldots,z_N) > 0$.
{\small
\begin{eqnarray}
  \ln \popd(y) & = & \ln\frac{p(z_N)p(z_{N-1}|z_N)\cdots p(z_1|z_2)p(y|z_1)}{p(z_N|y)p(z_{N-1}|z_N,y)\cdots p(z_1|z_2,y)} \nonumber\\
  \nonumber \\
  - \ln \popd(y) & = & \left\{\begin{array}{l} KL(p(z_N|y),\;p(z_N)) \\ \\ + \sum_{i=2}^N  \; E_{z_i|y}\; KL(p(z_{i-1}|z_i,y),\;p(z_{i-1}|z_i)) \\ \\ - E_{z_1|y}\;[\ln p(y|z_1)] \end{array}\right. \label{Markov-eq} \\
  \nonumber\\
  H(\popd(y)) & \leq & \left\{\begin{array}{l} E_y\;KL(p(z_N|y),\;p_\gen(z_N)) \\ \\ + \sum_{i=2}^N  \; E_{y,z_i}\; KL(p(z_{i-1}|z_i,y),\;p_\gen(z_{i-1}|z_i)) \\ \\ - \ln p_\gen(y|z_1) \end{array}\right. \;\;\;~\label{Markov-ELBO}
\end{eqnarray}
}This can be mapped to diffusion models by taking $z_i = z(i\Delta t)$ with $N\Delta t >> 1$.
VAE analyses of diffusion models typically construct a reverse-diffusion process by training generator models to minimize the upper bound in (\ref{Markov-ELBO}).
However, here we focus on (\ref{Markov-eq}) rather than (\ref{Markov-ELBO}).  For the limit $\Delta t \rightarrow 0$ section~\ref{sec:derivation} gives a direct
derivation of the following (without use of the Fokker-Planck Equation).
{\small
\begin{equation}
  p(z(t - \Delta t)|z(t),y) = {\cal N}\left(\begin{array}{l}z(t) + \frac{\Delta t(y - z(t))}{t}, \\ \\ \Delta t I\end{array}\right) \label{VAEa}
\end{equation}
\begin{equation}
  p(z(t - \Delta t)|z(t)) = {\cal N}\left(\begin{array}{l} z(t) + \frac{\Delta t(E[y|t,z(t)] - z(t))}{t}, \\ \\ \Delta t I\end{array}\right) \label{VAEb}
\end{equation}
}

Equation (\ref{VAEb}) defines a reverse-diffusion SDE which we can write as
{\small
\begin{equation}
  z(t - \Delta t) = \left\{\begin{array}{l} z(t) + \left(\frac{E[y|t,z(t)] - z(t)}{t}\right)\Delta t + \sqrt{\Delta t} \epsilon \\ \\\epsilon \sim {\cal N}(0,I)\end{array}\right. \label{reverse1}
\end{equation}
}For two Gaussian distributions with the same isotropic covariance we have
$$KL\left(\begin{array}{l}{\cal N}(\mu_1,\sigma^2 I), \\ {\cal N}(\mu_2,\sigma^2 I) \end{array}\right) = \frac{||u_1 -\mu_2||^2}{2\sigma^2}$$
Applying this to (\ref{VAEa}) and (\ref{VAEb}) we get
\begin{eqnarray}
  & & KL\left(\begin{array}{l}p(z(t-\Delta t)|z(t),y), \\p(z(t-\Delta t)|z(t))\end{array}\right) \nonumber\\
  \nonumber\\
  & = & \left(\frac{||y-E[y|t,z(t)]||^2\Delta t^2}{2t^2\Delta t}\right) \nonumber \\
  \nonumber \\
  & = & \left(\frac{||y-E[y|t,z(t)]||^2}{2t^2}\right) \Delta t \label{VAE}
\end{eqnarray}

We will consider (\ref{Markov-eq}) under the diffusion interpretation in the limit as $\Delta t \rightarrow 0$ and $N\Delta t \rightarrow \infty$.
Taking $N \Delta t \rightarrow \infty$ we have that the first term in (\ref{Markov-eq}) converges to zero.
As $\Delta t \rightarrow 0$ the second term in (\ref{Markov-eq}) diverges to infinity while the third term diverges to negative infinite allowing the equation to hold for all $\Delta t$.
To see why the third term diverges to negative infinity we can consider the one dimensional case where the population is uniformly distributed on the interval $[0,1]$.
For small $\Delta t$, and a given sample of $z(\Delta t)$ given $y$, the density $p(y|z(\Delta t))$ will be a Gaussian centered
at $z(\Delta t)$ with standard deviation $\sqrt{\Delta t}$.  As $\Delta t \rightarrow 0$ the central density will be proportional to $1/\sqrt{\Delta t}$.  The third term
then diverges to negative infinity at the rate of $\frac{1}{2}\ln \Delta t$.
To handle this difficulty we can fix a time $t_0 > 0$.
We then have that (\ref{Markov-eq}) and (\ref{VAE}) imply
{\small
\begin{equation}
  - \ln \popd(y) = \left\{\begin{array}{l}\int_{t_0}^\infty dt \;\; E_{z(t)|y}\;\left[\frac{||y - E[y|t,z(t)]||^2}{2t^2}\right] \\ \\ + \;E_{z(t_0)|y}[-\ln p(y|z(t_0))]\end{array}\right. \label{popd}
\end{equation}
}Taiking the expectation over $y$ of (\ref{popd}) we get
{\small
\begin{equation}
  H(\popd(y)) = \left\{\begin{array}{l}\int_{t_0}^\infty dt \;\; E_{y,z(t_0)}\;\left[\frac{||y - E[y|t,z(t_0)]||^2}{2t^2}\right] \\ \\ +\;H(y|z(t_0))\end{array}\right. \label{popb}
\end{equation}
}In (\ref{popb}), as $t_0 \rightarrow 0$ we have that the second term goes to negative infinity while the first term goes to positive infinity with both diverging at the rate of $\frac{1}{2}\ln t_0$.
This pathological behavior is due to the pathological nature of differential entropy.  A real number actually carries an infinite amount of information

In the continuous case it is better to work with mutual information as in Shannon's channel capacity theorem for continuous signals.
The mutual information $I(y,z(t_0)) = H(y) - H(y|z(t_0))$ is
{\small
  \begin{equation}
    I(y,z(t_0)) = \int_{t_0}^\infty dt\; E_{y,z(t)}\;\left[\frac{||y - E[y|t,z(t)]||^2}{2t^2}\right] \label{popd2}
  \end{equation}
}Note that as $t_0 \rightarrow 0$ we have that $I(y,z(t_0))$ approaches $I(y,y)$ which is infinite for continuous densities.

For images a natural choice of $t_0$ is the variance of camera noise.  A natural choice of discrete levels of $t$ for numerical integration seems to be uniform in $\ln t$.

\subsection{Estimating $E[y|t,z(t)]$}
  
Equations (\ref{reverse1}), (\ref{popd}) and (\ref{popd2}) are stated in terms of $E[y|t,z(t)]$.
We can train a network $\hat{y}(t,z)$ to estimate $E[y|t,z(t)]$ using
\begin{equation}
  \hat{y}^* = \argmin_{\hat{y}} E_{t,z(t)} \;(\hat{y}(t,z) - y)^2 \label{train1}
\end{equation}
In practice it is better to train a network on values of the same scale.  If the population values are scaled so as to have scale 1, then the scale of $z(t)$ is $\sqrt{1+t}$.
A constant-scale network can then take $z/\sqrt{1 + t}$ as an argument rather than $z$.
\begin{equation}
  \hat{y}^* = \argmin_{\hat{y}} E_{t,z(t)} \; (\hat{y}(t,z/\sqrt{1+t}) - y)^2 \label{train1}
\end{equation}
We then have
\begin{equation}
  E[y|t,z(t)] = \hat{y}^*(t,z/\sqrt{1+t}))
\end{equation}

\subsection{The Derivation of (\ref{VAEa}) and (\ref{VAEb})}
\label{sec:derivation}

We first condition the diffusion process on a particular value of $y$.
For $t > 0$ we can jointly draw $z(t)$ and $z(t+\Delta t)$ conditioned on $y$ as follows (taking $t > 0$ ensures that as $\Delta t \rightarrow 0$ we have $t >> \Delta t$).
\begin{eqnarray*}
  z(t) & = & y + \sqrt{t}\; \delta,\;\;\delta \sim {\cal N}(0,I) \\
  \\
  z(t+ \Delta t) & = & z(t) + \sqrt{\Delta t}\; \epsilon,\;\;\epsilon \sim {\cal N}(0,I)
\end{eqnarray*}
Next we additionally condition on the value of $z(t+\Delta t)$.
The choice of $y$, and $z(t + \Delta t)$ places a constraint on $\delta$ and $\epsilon$.
$$\sqrt{t} \delta + \sqrt{\Delta t} \epsilon = z(t + \Delta t) - y$$
This allows us to solve for $\delta$ as a function of $\epsilon$.
$$\sqrt{t} \delta + \sqrt{\Delta t} \epsilon = a\;\;\mathrm{for}\;\;{\color{red}a = z(t + \Delta t) - y}$$
$$\delta(\epsilon) = \frac{1}{\sqrt{t}}(a - \sqrt{\Delta t} \epsilon)$$
Given $\delta$ as a function of $\epsilon$ we can compute the density of $\epsilon$ given $y$ and $z(t + \Delta t)$.

{\small
\begin{eqnarray*}
  & & -  \ln p(\epsilon |z(t + \Delta t),y) \\
  \\
  & = & \frac{1}{2}||\delta(\epsilon)||^2 + \frac{1}{2}||\epsilon||^2 + C_1 \\
  \\
  & = & \frac{1}{2}\left|\left|\frac{1}{\sqrt{t}}(a - \sqrt{\Delta t} \epsilon)\right|\right|^2 + \frac{1}{2}||\epsilon||^2 + C_1 \\
  \\
  & = & \frac{\Delta t}{2t}||\epsilon||^2 -\frac{\sqrt{\Delta t} a^\top \epsilon}{t} + \frac{1}{2}||\epsilon||^2 + C_2
  \\
  & = & \eta\left(||\epsilon||^2 -\frac{\sqrt{\Delta t} a^\top \epsilon}{\eta t}\right) + C_2 \;\;{\color{red} \eta = \frac{\Delta t}{2t} + \frac{1}{2} \approx \frac{1}{2}}\\
  \\
  & = & \eta\left|\left|\epsilon -\frac{\sqrt{\Delta t} a}{2\eta t}\right|\right|^2 + C_3 \\
  \\
  & \approx & \frac{1}{2} \left|\left|\epsilon -\frac{\sqrt{\Delta t} a}{t}\right|\right|^2 + C_3 \\
\end{eqnarray*}
}
This gives
$$p(\epsilon|z(t + \Delta t),y) = {\cal N}\left(\frac{\sqrt{\Delta t}\;(z(t+\Delta t) - y)}{t},\;I\right)$$
We have $z(t) = z(t +\Delta t) - \sqrt{\Delta t}\epsilon$ which now gives
\begin{eqnarray*}
  & & p(z(t)|z(t + \Delta t),y) \\
  & = & {\cal N}\left(z(t + \Delta t) + \frac{\Delta t(y - z(t+\Delta t))}{t},\;\Delta t I\right)
\end{eqnarray*}
This is (\ref{VAEa}).
This implies
\begin{eqnarray*}
  & & E[z(t)|z(t+ \Delta t)] \\
  & = & z(t + \Delta t) + \frac{\Delta t(E[y|t,z(t+\Delta t)] - z(t + \Delta t))}{t}
\end{eqnarray*}
We also have
$$p(z(t)|z(t+ \Delta t)) = E_y[p(z(t)|z(t + \Delta t),y)]$$
This implies that $p(z(t)|z(t+ \Delta t))$ is a mixture of Gaussians and hence is Gaussian.
The mean of the distribution equals $z(t + \Delta t)$ plus a term proportional to $\Delta t$.  The standard deviation (the square root of the variance)
is proportional to $\sqrt{\Delta t}$ which, in the limit of small $\Delta t$, is infinitely larger than $\Delta t$.
This implies that, to leading order in $\Delta t$, the variance remains $\Delta t I$.  So we now have
\begin{eqnarray*}
  & & p(z(t)|z(t+\Delta t)) \\
  & = & {\cal N}\left(\left(\begin{array}{l}z(t + \Delta t) \\
      + \frac{\Delta t(E[y|t,z(t+ \Delta t)] - z(t + \Delta t))}{t}\end{array}\right),
  \Delta t I\right)
\end{eqnarray*}
This is (\ref{VAEb}).

\section{The Fokker-Planck Analysis (Score Matching)}

A general stochastic differential equation can be formulated as the limit as $\Delta t \rightarrow 0$
of a discrete-time difference equation
\begin{equation}
  z(t+\Delta t) = \left\{\begin{array}{l} z(t) + v(z(t),t)\Delta t + \sqrt{\Delta t}\; \epsilon \\ \\ \epsilon \sim {\cal N}{(0,\Sigma(z(t),t))}\end{array}\right. \label{SDE}
\end{equation}
The diffusion process is the special case given by (\ref{forward}) which we repeat here.
$$z(t + \Delta t) = z(t) + \sqrt{\Delta t}\;\epsilon,\;\;\;\epsilon \sim {\cal N}(0,I)\;\;\;\;\;\;\; \mbox{(\ref{forward})}$$

The Fokker-Planck equation governs the time evolution of the probability density $p_t(z)$ for an SDE (\ref{SDE}).
\begin{equation}
  \frac{\partial p_t(z)}{\partial t} = - \nabla \cdot\left(\begin{array}{l}v(z(t),t)p_t(z) \\ \\ - \frac{1}{2}\Sigma(z(t),t) \nabla_z p_t(z)\end{array}\right) \label{FP}
\end{equation}

For the special case of (\ref{forward}) the Fokker-Planck equation becomes
\begin{equation}
\frac{\partial p_t(z)}{\partial t} = - \nabla \cdot\left(-\frac{1}{2}\nabla_z p_t(z)\right) \label{FP2}
\end{equation}

One can gain intuition for this equation by considering the density of perfume in the air over time.  Perfume expands into air by a diffusion process.
There is a diffusion-flow of the perfume.  The perfume flows in a direction from higher concentration to lower concentration. For the diffusion process defined here
the diffusion flow vector at ``position'' $z$ and time $t$ is $-(1/2) \nabla _z p_t(z)$.  The time derivative of the concentration at a particular position $z$ and time $t$
is determined by the difference between the flow-out and the flow-in for a little ball about $z$.  For a flow vector $F$ the flow-out minus flow-in (per unit volume)
is given by the divergence $\nabla \cdot F$.  The flow-in minus the flow-out is $-\nabla \cdot F$.

We can rewrite (\ref{FP2}) as

\begin{equation}
\frac{\partial p_t(z)}{\partial t} = - \nabla \cdot\left[\left(-\frac{1}{2}\nabla_z \ln p_t(z)\right)p_t(z)\right] \label{FP3}
\end{equation}

Comparing this with the general case of the Fokker-Planck equation (\ref{FP}) we can see that (\ref{FP3}) can be interpreted as the case of a velocity vector with zero noise.
We have that the time evolution of $p_t(z)$ under (\ref{forward}) is the same as that given by (\ref{FP2}) which is the same as that given by (\ref{FP3})
which, in turn, is the same as the time evolution of $p_t(z)$ under a deterministic velocity field given by
\begin{equation}
  \frac{dz}{dt} = - \frac{1}{2}\nabla_z \ln p_t(z) \label{ODE}
\end{equation}
The time-reversal of this deterministic differential equation gives a deterministic reverse-diffusion process.

For any density $p(z)$ the gradient vector $\nabla_z\;\ln p(z)$ is called the score function.
For the diffusion process we can solve for the score function as follows.
{\small
\begin{eqnarray}
  p_t(z) & = & E_y \;p_t(z|y) \nonumber \\
  \nonumber \\
  & = & \;E_y\;\frac{1}{Z(t)} e^{-\frac{||z-y||^2}{2t}} \nonumber \\
  \nonumber \\
  \nabla_z p_t(z) & = & E_y \;p_t(z|y)\; (y-z)/t \nonumber \\
  \nonumber \\
  & = & E_{z,y}\; (y-z)/t \nonumber \\
  \nonumber \\
   & = & p_t(z)\;E_{y|z}\;(y-z)/t \nonumber \\
  \nonumber \\
  & = & p_t(z)\; (E[y|t,z]-z)/t \nonumber \\
  \nonumber \\
  \nabla_z \ln p_t(z) & = & \frac{\nabla_z p_t(z)}{p_t(z)} \;\;= \frac{E[y|t,z]-z}{t} \label{Score}
\end{eqnarray}
}
The deterministic reverse-diffusion process (\ref{ODE}) can now be written as
\begin{equation}
  z(t - \Delta t) = z(t) + \left(\frac{E[y|t,z(t)] - z(t)}{2t}\right)\Delta t \label{reverse2}
\end{equation}
It is useful to compare (\ref{reverse2}) with the reverse-diffusion SDE (\ref{reverse1}) derived by direct analysis.  The reverse-diffusion SDE (\ref{reverse1}) has exactly twice the linear term as the deterministic reverse-diffusion
process (\ref{reverse2}). Intuitively, noise broadens the distribution whether it is applied forward or backward in time. To reverse the process we can apply
the noise for $\Delta t$, which ``diffuses'' the distribution by $\Delta t$ but then reverse that diffusion by doubling the reverse-diffusion term.
More rigorously, if we define $t'$ = $-t$ then (\ref{reverse1}) can be written as
{\small
\begin{eqnarray*}
  z(t' + \Delta t') & = & z(t') + \left(\frac{E[y|t,z(t)] - z(t)}{t}\right)\Delta t' + \sqrt{\Delta t}\; \epsilon \\
  \\
  & & \epsilon \sim {\cal N}(0,I)
\end{eqnarray*}
}
Applying the Fokker-Planck equation to this yields
\begin{eqnarray*}
  \frac{\partial p_t(z)}{\partial t'} & = & - \nabla \cdot\left((\nabla_z \ln p_t(z))p_t(z)) - \frac{1}{2}\nabla_z\;p_t(z)\right) \\
  \\
  & = & - \nabla \cdot \left[\left(\frac{1}{2} \nabla_z \ln p_t(z)\right)p_z(t)\right]
\end{eqnarray*}
This last equation corresponds to the time-reversal of deterministic differential equation (\ref{ODE}).  Hence the Fokker-Planck analysis of (\ref{reverse1}) also gives that
(\ref{reverse1}) reverses the time evolution of the density $p_t(z)$.

We can generalize the reverse-diffusion process to include any degree of stochasticity.  For any $\lambda \geq 0$ we can use the SDE
\begin{equation}
  z(t - \Delta t)  =   \left\{\begin{array}{l} z(t) \\ + \frac{1+\lambda}{2}\left(\frac{E[y|t,z(t)] - z(t)}{t}\right)\Delta t \\ + \lambda\epsilon \sqrt{\Delta t} \\ \\ \epsilon \sim {\cal N}(0,I)\end{array}\right. \label{INV2}
\end{equation}
This provides a potentially useful hyper-parameter for emperical optimization.

\section{Conclusions}

Both the VAE analysis and the Fokker-Planck analysis yield reverse-diffusion processes for sampling from a population. Fokker-Planck
yields reverse-diffusion processes parameterized by a reverse-diffusion noise level.
The VAE analysis yields likelihood formulas. In contrast to some discussions in the literature, Langevin dynamics and simulated annealing seem
unrelated to reverse-diffusion.

\section{Attributions}

The modern literature on sampling by reverse-diffusion originates in \cite{Diffusion}.
They
employ a VAE analysis in which generator distributions are trained by optimizing the variational bound (\ref{Markov-ELBO}).
More effective methods for optimizing the variational bound were given in \cite{DDPM} and a large number of empirical refinements to this approach have appeared in the literature.

The modern literature on the Fokker-Planck analysis of diffusion models (score matching) originates in \cite{Score}.
The reverse-diffusion SDE derived from the Fokker-Planck equation appears in \cite{Anderson} which is cited by Song et al.
Song et al. derive the ODE (\ref{ODE}).
The closed form expression for the score function (\ref{Score}) can be viewed as a special case of Tweedie's formula which first appears in \cite{TweedieRobbins}.
Although Song et al. mention Tweedie's formula, they do not use (\ref{Score}) in their score-matching algorithm.
Instead they optimize a score-matching objective from \cite{ScoreMatching}.  The use of (\ref{Score}) in the reverse-diffusion ODE and SDE can be found in \cite{Karras}.

Equation (\ref{popd2}) is equivalent to the information minimum mean squared error relation (I-MMSE) \cite{I-MMSE}.  A discussion of the relationship between I-MMSE and diffusion models, and a form of equation (\ref{popd}), can be found in \cite{InfoDiff}.

%\bibliography{unification}
%\bibliographystyle{icml2022}

\end{document}